# Hybrid Variational/Gibbs Collapsed Inference in Topic Models

**Max Welling***
Dept. Computer Science
University of California Irvine
Irvine, CA, USA
*welling@ics.uci.edu*

**Yee Whye Teh**
Gatsby Computational Neuroscience Unit
University College London
London, UK
*ywteh@gatsby.ucl.ac.uk*

**Bert Kappen**
Dept. Biophysics
Radboud University Nijmegen
Nijmegen, The Netherlands
*b.kappen@science.ru.nl*

## Abstract

Variational Bayesian inference and (collapsed) Gibbs sampling are the two important classes of inference algorithms for Bayesian networks. Both have their advantages and disadvantages: collapsed Gibbs sampling is unbiased but is also inefficient for large count values and requires averaging over many samples to reduce variance. On the other hand, variational Bayesian inference is efficient and accurate for large count values but suffers from bias for small counts. We propose a hybrid algorithm that combines the best of both worlds: it samples very small counts and applies variational updates to large counts. This hybridization is shown to significantly improve test-set perplexity relative to variational inference at no computational cost.

## 1 Introduction

Bayesian networks (BNs) represent an important modeling tool in the field of artificial intelligence and machine learning (Heckerman, 1999). In particular the subclass of BNs known as "topic models" is receiving increasing attention due to its success in modeling text as a bag-of-words and images as a bag-of-features (Blei et al., 2003). Unlike most applications of Bayesian networks, we will be interested in "Bayesian Bayesian networks" where we also treat the conditional probability tables (CPTs) as random variables (RVs). The key computational challenge for these models is inference, namely estimating the posterior distribution over both parameters and hidden variables, and ultimately estimating predictive probabilities and the marginal log-likelihood (or evidence).

It has been argued theoretically (Castella & Robert, 1996) and observed empirically in topic models (Griffiths & Steyvers, 2002; Buntine, 2002) that Gibbs sampling in a collapsed state space where the CPTs have been marginalized out leads to efficient inference. It is expected that this also holds more generally true for discrete Bayesian networks. Variational Bayesian (VB) approximations have also been applied to topic models (Blei et al., 2003) but predictive probability results have consistently been inferior to collapsed Gibbs sampling (CGS). More recently, variational approximations have been extended to operate in the same collapsed state space of CGS (Teh et al., 2006; Teh et al., 2008). These collapsed variational Bayesian (CVB) inference algorithms improve upon VB but still lag behind CGS.

In this paper we will propose a hybrid inference scheme that combines CGS with VB approximations. The idea is to split all data-cases into two sets, the ones that will be treated variationally and the ones that will be treated through sampling. The two approximations interact in a consistent manner in the sense that both VB and CGS updates are derived from a single objective function. The advantage of the VB updates is that they scale better computationally. We show empirically that hybrid algorithms achieve almost the same accuracy as CGS because they are only applied where they are expected to work well. The algorithm can be seen to trade off bias with variance in a flexible and tunable manner.

## 2 Topic Models as Bayesian Networks

We will assume that all visible and hidden variables are discrete. However, we expect the results to hold more generally for models in the exponentially family. We will first develop the theory for latent Dirichlet allocation (LDA) (Blei et al., 2003) and later generalize the results to Bayesian networks. To facilitate the transition from LDA to BNs we will treat LDA in a slightly unconventional way by using a single index $i$ that runs over all words in all documents (in contrast to the index $ij$ which is conventional for LDA), see Fig.1. LDA is equivalent to the Bayesian network $d_i \rightarrow z_i \rightarrow x_i$, where nodes $x_i = w$ (word-type) and $d_i = j$

*On Sabbatical at Radboud University, Netherlands, Department of Biophysics.

(document label) have been observed. The topic variable is indicated by $z_i = k$. In the following we will also use the indicator variables $X_{iw} = \mathbb{I}[x_i = w]$, $D_{ij} = \mathbb{I}[d_i = j]$ and $Z_{ik} = \mathbb{I}[z_i = k]$.

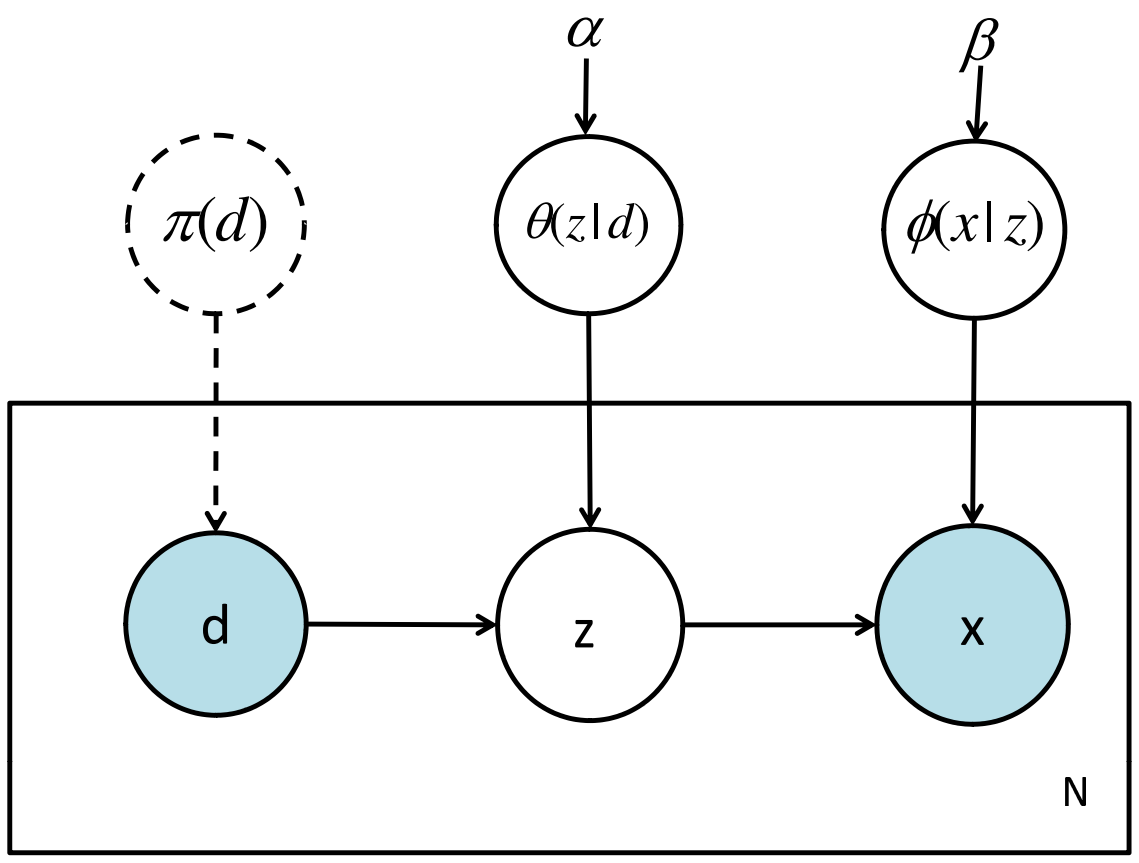

Figure 1: LDA depicted as a standard Bayesian network. $N$ runs over all word-tokens and for each token, both the word-type $x$ and the document label $j$ are observed. The topic variable $z$ is hidden.

The joint distribution is given as the product of the probability of each variable given its parents, $p(x_i = w|z_i = k) = \phi_{wk}$, $p(z_i = k|d_i = j) = \theta_{kj}$ and $p(d_i = j) = \pi_j$. The last term decouples from the rest of the model because $d_i$ is always observed. For each of these CPTs we place a Dirichlet prior, $\phi_{:,k} \sim \mathcal{D}(\beta)$ and $\theta_{:,j} \sim \mathcal{D}(\alpha_:)$. Note that we have chosen a scalar strength parameter $\beta$ (i.e. a symmetric Dirichlet prior) for $\phi$ but vector valued strength parameters $\alpha_: = \{\alpha_k\}$ for $\theta$.

Next we integrate out the CPTs. Since the Dirichlet priors are conjugate to the discrete conditional probabilities of the Bayesian network, this is possible analytically, resulting in the following expression,

$$p(\{z_i, x_i\}|\{d_i\}) \propto \frac{\prod_{wk} \Gamma(N_{wk} + \beta)}{\prod_k \Gamma(N_k + W\beta)} \prod_{jk} \Gamma(N_{kj} + \alpha_k) \quad (1)$$

where $W$ is the total number of different word types (the vocabulary size), and the counts are defined as follows:

$$N_{wkj} = \sum_i X_{iw} Z_{ik} D_{ij}, \quad (2)$$

$N_{wk} = \sum_j N_{wkj}$, $N_{kj} = \sum_w N_{wkj}$ and $N_k = \sum_{wj} N_{wkj}$. Note that $N_{wkj}$ are subject to the observational constraints $\sum_k N_{wkj} = \hat{N}_{wj}$ given by the training corpus.

# 3 Standard Variational Bayes

In the variational Bayesian framework we approximate the posterior distribution with a factorized one:

$$p(\{z_i, \phi, \theta\}|\{x_i, d_i\}) \approx \prod_k Q(\phi_{:,k}|\xi_{:,k}) \prod_j Q(\theta_{:,j}|\zeta_{:,j}) \prod_i Q(z_i) \quad (3)$$

Solving for $Q(\phi_{:,k})$ and $Q(\theta_{:,j})$, we find that they are Dirichlet as well, with parameters,

$$\xi_{wk} = \beta + \bar{N}_{wk}, \qquad \zeta_{kj} = \alpha_k + \bar{N}_{kj} \quad (4)$$

where $\bar{N}_{wk} = \sum_i Q_{ik} X_{iw}$, $\bar{N}_{kj} = \sum_i Q_{ik} D_{ij}$ and $Q_{ik} = Q(z_i = k)$. After plugging these updates back into the update for $Q(z_i)$ we find,

$$Q_{ik} \propto \frac{\exp(\psi(\bar{N}_{x_i k} + \beta))}{\exp(\psi(\bar{N}_k + W\beta))} \exp(\psi(\bar{N}_{kd_i} + \alpha_k)) \quad (5)$$

where $\psi(\cdot) = \frac{\partial}{\partial x} \log \Gamma(x)$ is the derivative of the log-gamma function.

For future reference we shall now derive the same result in a different manner to which we shall refer as "standard variational Bayes" (SVB) in the following. We first marginalize out the CPTs and consider a variational lower bound on the evidence with a variational distribution $Q(\{z_i\})$ (without assuming factorization). Lemma 1 in the appendix shows that the expression for the log-probability $\log p(\{z_i, x_i\}|\{d_i\})$ is a convex function of $\{N_{wjk}\}$. This allows us to move the average over $Q(\{z_i\})$ inside the expression for the log-probability, at the cost of introducing a further lower bound on the evidence. The resulting expression is now precisely the logarithm of Eqn.1 with the counts $N_{wkj}$ replaced by average counts $\bar{N}_{wkj}$,

$$\bar{N}_{wkj} = \mathbb{E}[N_{wkj}]_Q = \hat{N}_{wj} Q_{k|wj} \quad (6)$$

in terms of which we define

$$\bar{N}_{wk} = \mathbb{E}[N_{wk}]_Q = \sum_j \hat{N}_{wj} Q_{k|wj} \quad (7)$$

$$\bar{N}_{kj} = \mathbb{E}[N_{kj}]_Q = \sum_w \hat{N}_{wj} Q_{k|wj} \quad (8)$$

$$\bar{N}_k = \mathbb{E}[N_k]_Q = \sum_{wj} \hat{N}_{wj} Q_{k|wj} \quad (9)$$

where $\hat{N}_{wj}$ are the observed word-document counts. To understand the definition of $Q_{k|wj}$ we note that $Q(z_i)$ for all $i$'s with the same values of $x_i = w$, $d_i = j$ are equal, so without loss of generality we can use a single set of parameters $Q_{k|wj} = Q(z_i = k)$ for all data-cases $i$ which share the same observed labels $w, j$.

The final step is to variationally bound this expression once more, using again the fact that the sum of log-gamma factors is a convex function,

$$F(\bar{N}^*_{wkj}) \geq F(\bar{N}_{wjk}) + \sum_{wkj} \nabla_{wkj} F(\bar{N}_{wkj})(\bar{N}^*_{wjk} - \bar{N}_{wkj}) \tag{10}$$

where $\bar{N}_{wkj}$ is held fixed. Recalling definition 6 and taking derivatives w.r.t $Q_{k|wj}$ gives the following update,

$$Q_{k|wj} \propto \frac{\exp(\psi(\bar{N}_{wk} + \beta))}{\exp(\psi(\bar{N}_k + W\beta))} \exp(\psi(\bar{N}_{jk} + \alpha_k)) \tag{11}$$

The factorization follows directly from the update and is a result of Jensen's inequality. We can alternatively arrive at Eqn.11 without assuming the second bound of Eqn.10 by assuming that $Q(\{z_i\})$ factorizes and equating its derivatives to 0. However, this does not guarantee convergence as $Q_{k|wj}$ now also appears on the RHS of Eqn.11. Note that Eqn.11 is equivalent but looks subtly different from Eqn.5 in that it avoids updating variational distributions for data-cases $i$ with the same labels $w, j$. As a result it scales more favorably as the number of unique word-document pairs in the training corpus rather than the total number of word tokens.

## 4 Collapsed Gibbs Sampling

An alternative to variational inference is collapsed Gibbs sampling where one samples each $z_i$ in turn, given the values of the remaining $z_{\neg i}$. The conditional probabilities are easily calculated from Eqn.1,

$$p(z_i = k | \mathbf{z}_{\neg i}, \mathbf{x}, \mathbf{d}) \propto \frac{(N^{\neg i}_{wk} + \beta)}{(N^{\neg i}_k + W\beta)} (N^{\neg i}_{jk} + \alpha_k) \tag{12}$$

where the superscript $^{\neg i}$ denotes that data-case $i$ has been removed from the count and we have assumed that $x_i = w$ and $d_i = j$.

Given samples at equilibrium one can obtain unbiased estimates of quantities of interest. The trade-off is that one needs to average over many samples to reduce the effects of sampling noise. Computationally, CGS scales as $\mathcal{O}(NK)$ in time and $\mathcal{O}(N)$ in space, where $N$ is the total number of words in the corpus and $K$ is the number of topics. This in contrast to the SVB updates in the previous section for which both time and space scale as $\mathcal{O}(MK)$ with $M$ the number of unique word-document pairs.

In the following section we will derive a principled hybridization of SVB and CGS that can be viewed as a tunable trade-off between bias, variance and computational efficiency.

## 5 The Hybrid SVB/CGS Algorithm

The high level justification for a hybrid algorithm is the intuition that the evidence only depends on count arrays, $N_{wk}$, $N_{kj}$ and $N_k$, which are sums of assignment variables $N_{wkj} = \sum_i X_{iw} Z_{ik} D_{ij}$ where only $Z$ is random. Also the Rao-Blackwellised estimate of the predictive distribution is a function of these counts,

$$p(x^* = w | \{x_i, z_i, d_i\}, d^* = j) = \sum_k \frac{N_{wk} + \beta}{N_k + W\beta} \frac{N_{kj} + \alpha_k}{N_j + \sum_k \alpha_k} \tag{13}$$

The central limit theorem tells us that sums of random variables tend to concentrate and behave like a normal distribution under certain conditions. Moreover, the variance/covariance of the predictive distribution is expected to scale with $1/n$, where $n$ is the number of data-cases that contribute to the sum. We expect that variational approximations work well for large counts.

These insights make it natural to split the dataset into two subsets, one subset $\mathcal{S}^{\text{VB}}$ to which we will apply the VB approximation and the complement $\mathcal{S}^{\text{CG}}$ to which we shall apply collapsed Gibbs sampling. In practice we have chosen these sets to be:

$$\mathcal{S}^{\text{VB}} = \{i | \hat{N}_{x_i, d_i} > r\}, \qquad \mathcal{S}^{\text{CG}} = \{i | \hat{N}_{x_i, d_i} \leq r\} \tag{14}$$

In the experiments below we have chosen $r = 1$. Although we do not expect that central limit tendencies apply to counts smaller than about 10, we have chosen this extreme setting to convey an important conclusion, namely that the counts with value $\hat{N}_{wj} = 1$ already explain much of the difference between VB and CGS algorithms.

We will assume the following factorization for the variational posterior, $Q = Q^{\text{VB}} Q^{\text{CG}}$. Moreover, in the derivation below it will follow that $Q^{\text{VB}}$ becomes factorized as well, i.e. $Q^{\text{VB}} = \prod_i Q^{\text{VB}}_i$.

The evidence of the collapsed distribution under these assumptions reads,

$$\mathcal{E} = \mathcal{H}(Q^{\text{VB}}) + \mathcal{H}(Q^{\text{CG}}) + \sum_{\mathbf{z}^{\text{VB}}, \mathbf{z}^{\text{CG}}} Q(\mathbf{z}^{\text{VB}}) Q(\mathbf{z}^{\text{CG}}) \log P(\mathbf{z}^{\text{VB}}, \mathbf{z}^{\text{CG}}, \mathbf{x} | \mathbf{d}) \tag{15}$$

Analogous to section 3 we apply Jensen's inequality in order to bring the average over $Q^{\text{VB}}$ inside the convex function $\log P(\mathbf{z}^{\text{VB}}, \mathbf{z}^{\text{CG}}, \mathbf{x})$, resulting in an expression which we shall denote as $\log P(Q^{\text{VB}}, \mathbf{z}^{\text{CG}}, \mathbf{x} | \mathbf{d})$ for obvious reasons. The log-probability only depends on counts, so by bringing the average inside we find that $\log P(Q^{\text{VB}}, \mathbf{z}^{\text{CG}}, \mathbf{x}, \mathbf{d})$ depends on the quantities,

$$\mathbb{E}[N_{wkj}]_{Q^{\text{VB}}} = \hat{N}^{\text{VB}}_{wj} Q^{\text{VB}}_{k|wj} + \sum_{i \in \mathcal{S}^{\text{CG}}} z_{ik} x_{iw} d_{ij} \tag{16}$$

where $\hat{N}^{\text{VB}}_{wj}$ are the observed counts for word-type $w$ and document $j$ which are in the set $\mathcal{S}^{\text{VB}}$. In terms of this we further need,

$$\mathbb{E}[N_{wk}]_{Q^{\text{VB}}} = \sum_j \hat{N}^{\text{VB}}_{wj} Q^{\text{VB}}_{k|wj} + \sum_{i \in \mathcal{S}^{\text{CG}}} z_{ik} x_{iw} \quad (17)$$

$$\mathbb{E}[N_{jk}]_{Q^{\text{VB}}} = \sum_w \hat{N}^{\text{VB}}_{wj} Q^{\text{VB}}_{k|wj} + \sum_{i \in \mathcal{S}^{\text{CG}}} z_{ik} d_{ij} \quad (18)$$

$$\mathbb{E}[N_k]_{Q^{\text{VB}}} = \sum_{wj} \hat{N}^{\text{VB}}_{wj} Q^{\text{VB}}_{k|wj} + \sum_{i \in \mathcal{S}^{\text{CG}}} z_{ik} \quad (19)$$

These expressions elegantly split the counts into a part described through a non-random mean field plus a sum over the remaining random variables that represent the fluctuations.

Thus, after applying Jensen's inequality we end up with the following lower bound on the evidence,

$$\mathcal{B} = \mathcal{H}(Q^{\text{VB}}) + \mathcal{H}(Q^{\text{CG}}) + \sum_{\mathbf{z}^{\text{CG}}} Q(\mathbf{z}^{\text{CG}}) \log P(Q^{\text{VB}}, \mathbf{z}^{\text{CG}}, \mathbf{x}|\mathbf{d}) \leq \mathcal{E} \quad (20)$$

Now let's assume we have drawn a sample from $Q^{\text{CG}}$, which we will denote with $\mathbf{z}^{\text{CG}}_s$. Furthermore, denote with $\bar{N}^s_{wjk}$ the value of $\mathbb{E}[N_{wjk}]_{Q^{\text{VB}}}$ evaluated at $\mathbf{z}^{\text{CG}}_s$ (see Eqn.16). Given this sample we bound again through linearization (see Eqn.10) which results in the following update for $Q^{\text{VB}}$,

$$Q^{\text{VB}}_{k|wj} \propto \frac{\exp(\psi(\bar{N}^s_{wk} + \beta))}{\exp(\psi(\bar{N}^s_k + W\beta))} \exp(\psi(\bar{N}^s_{jk} + \alpha_k)) \quad (21)$$

In case we decide to use more than 1 sample we replace the expressions $\psi(\cdot) \rightarrow \langle\psi(\cdot)\rangle$, where the brackets $\langle\cdot\rangle$ denote taking the sample average.

The update for $Q^{\text{CG}}$ will be sample-based. We first compute the variational update for $Q^{\text{CG}}$,

$$Q^{\text{CG}} \propto P(Q^{\text{VB}}, \mathbf{z}^{\text{CG}}, \mathbf{x}|\mathbf{d}) \quad (22)$$

and subsequently draw samples from it. On closer inspection we see that this distribution is identical to the collapsed distribution $p(\mathbf{x}, \mathbf{z}|\mathbf{d})$, but over fewer data-cases, namely those in the set $\mathcal{S}^{\text{CG}}$, and with new effective hyper-parameters given by,

$$\alpha'_{jk} = \alpha_k + \sum_w \hat{N}^{\text{VB}}_{wj} Q^{\text{VB}}_{k|wj} \quad (23)$$

$$\beta'_{wk} = \beta + \sum_j \hat{N}^{\text{VB}}_{wj} Q^{\text{VB}}_{k|wj} \quad (24)$$

Hence, we can apply standard collapsed Gibbs sampling to draw from $Q^{\text{CG}}$,

$$p(\mathbf{z}^{\text{CG}}_i = k|\mathbf{z}^{\text{CG}}_{\neg i}, \mathbf{x}, \mathbf{d}) \propto \frac{(\bar{N}^{\neg i,s}_{wk} + \beta)}{(\bar{N}^{\neg i,s}_k + W\beta)} (\bar{N}^{\neg i,s}_{jk} + \alpha_k) \quad (25)$$

These updates converge in expectation and stochastically maximize the expression for the bound on the evidence. In theory one should draw infinitely many samples from $Q^{\text{CG}}$ to guarantee convergence. In practice however we have obtained very good results with drawing only a single sample before proceeding to the VB update.

It is also possible to infer the hyper-parameters $\{\alpha_k, \beta\}$ by either using sampling (Teh et al., 2004) or maximization (Minka, 2000). We refer to those papers for further details.

### 5.1 Extension To Collapsed Variational LDA

In (Teh et al., 2006) an improved variational approximation was proposed that operates in the same collapsed space as collapsed Gibbs sampling. This algorithm uses a factorized distribution $Q^{\text{CVB}} = \prod_i Q(z_i)$ but does not move this inside the log-probability as we did for SVB. Instead, it evaluates the necessary averages in the updates by assuming that the counts behave approximately normal and using a second order Taylor expansion around the mean. It was shown that including this second order information improves the standard VB approximation considerably.

This algorithm is also straightforwardly hybridized with collapsed Gibbs sampling by simply replacing the SVB updates with CVB updates in the hybrid SVB/CGS algorithm and using the same definitions for the counts as in Eqn.16. We call this algorithm CVB/CGS.

## 6 Extension to Bayesian Networks

The extension to collapsed Bayesian networks is relatively straightforward. First let's generalize SVB to Bayes nets. The derivation which includes variational distributions for the CPTs can be found in (Beal, 2003). We follow an alternative derivation that facilitates the transition to the hybrid SVB/CGS algorithm. We first collapse the state space by marginalizing out the CPTs. This results in an expression for the evidence that consists of products of factors, where each factor is a ratio of gamma-functions and the factors follow the structure of the original Bayes net. For instance, consider a hidden variable $z$ which can assume state values $k$ with two parents $u, v$ which take values $l, m$ respectively, see Fig.2. The factor associated with this family that will appear in the joint collapsed probability distribution of the BN is given by

$$F(\{z_i, u_i, v_i\}) = \prod_{lm} \left[ \frac{\Gamma(\sum_k \alpha_k)}{\Gamma(N_{lm} + \sum_k \alpha_k)} \prod_k \left[ \frac{\Gamma(N_{klm} + \alpha_k)}{\Gamma(\alpha_k)} \right] \right] \quad (26)$$

The complete joint (collapsed) probability distribution is given by a product of such factors, one for each family in the BN. This implies that the collapsed distribution inherits the graphical structure of the original BN, in particular its treewidth, implying that collapsed inference using SVB is equally expensive as ordinary inference (for given CPTs) in the original BN.

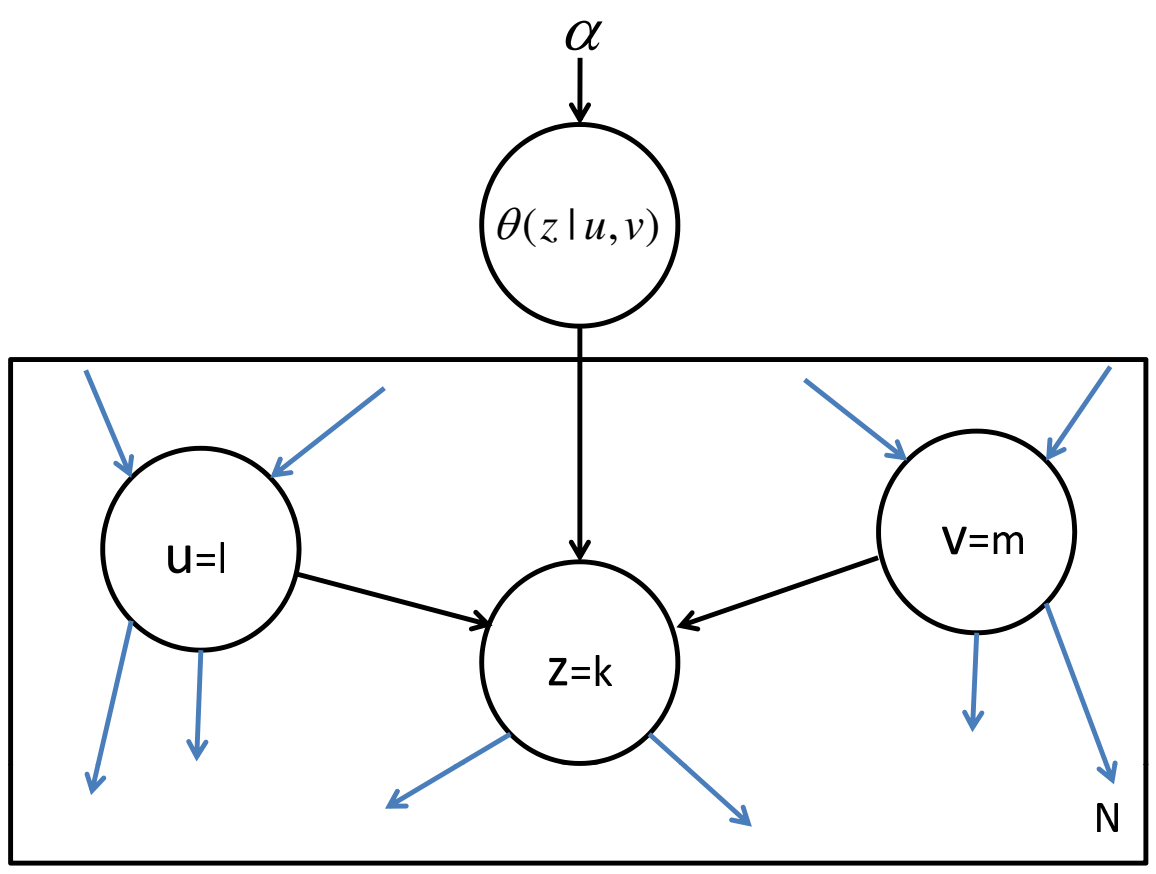

Figure 2: Family inside a BN consisting of two parents and one child.

Next, we move the average over the variational distribution $Q(\{\mathbf{z}_i\})$ inside the gamma-factors. This again produces a bound on the evidence. We then linearize the log-gamma factors around the current value of $Q$. Ignoring constant factors this results in the following terms for the family above,

$$G(Q, Q^*) = \text{terms for other families}+ \tag{27}$$
$$\sum_{klm}\left(\psi(\bar{N}_{klm}+\alpha_k)-\psi(\bar{N}_{lm}+\sum_k\alpha_k)\right)\bar{N}^*_{klm}$$

We recall again that $\bar{N}_{klm} = \sum_{w..}\hat{N}_{w..}Q_{klm|w..}$ where $w..$ represents all observed labels and $Q_{klm|w..}$ is the posterior marginal distribution over the variables $z, u, v$ given observations.

All of these factors are local in the cliques of the original Bayes net. Hence, the update for $Q$ becomes proportional to the product of local factors of the form,

$$F_{klm} = \frac{\exp(\psi(\bar{N}_{klm}+\alpha_k))}{\exp(\psi(\bar{N}_{lm}+\sum_k\alpha_k))} \tag{28}$$

As a result we can recompute new values for the local posterior marginals by running belief propagation over the junction tree associated with the original BN. Since $Q_{klm|w..}$ depends on $w..$, this has to be done for every combination of observed labels that has been observed at least once in the data (i.e. $\hat{N}_{w..} > 0$). It's interesting that the algorithm can thus be interpreted as an iterative belief propagation algorithm on a temporary graphical model where the potentials change from one iteration to the next. It bears a strong resemblance with iterative proportional fitting in which scaling updates enforce the constraints and alternate with message passing. In this interpretation, one could view the relations $\sum_{klm} N_{klm|w..} = \hat{N}_{w..}$ (which are equivalent to normalization of the $Q_{klm|w..}$) to be the constraints.

An alternative to VB is collapsed Gibbs sampling. Here one updates all hidden variables for a single data-item. One first removes the data-case from the pool and computes the expression for $p(\mathbf{z}_i|\mathbf{z}^{\neg i}, \mathbf{x}, \mathbf{d})$ over all hidden variables $\mathbf{z}$ in the Bayes net. This expression also factorizes according to the structure of the BN but the factors are now given by,

$$F'_{klm} = \frac{N^{\neg i}_{klm}+\alpha_k}{N^{\neg i}_{lm}+\sum_k\alpha_k} \tag{29}$$

One can draw samples by starting out at the leafs of the junction tree and computing distributions for the current node conditioned on upstream variables but marginalizing over all downstream variables. Given these variables we can then run an ancestral sampling pass outwards, back to the leaf nodes. This algorithm is an extension of the forward-filtering-backwards-sampling (FFBS) algorithm proposed in (Scott, 2002) for HMMs.

The derivation for the hybrid algorithm goes along similar lines as for LDA. We split all data-cases into two subsets, $\mathcal{S}^{\text{VB}}$ and $\mathcal{S}^{\text{CG}}$. We use Jensen's inequality to move the variational distribution $Q^{\text{VB}}$ inside the log-gamma functions. Finally, we derive updates for $Q^{\text{VB}}$ through the linearization trick. The final algorithm thus rotates over all data-cases, running either BP on the associated junction tree if the data-case is in $\mathcal{S}^{\text{VB}}$ (updating the $Q^{\text{VB}}$ for all families of the BNs) or the FFBS algorithm if the data-case is in $\mathcal{S}^{\text{CG}}$. The expressions for the counts are always the analogs of those in Eqn.16.

## 7 Experiments

We report results on two datasets: 1) "KOS", which is harvested from a lefty blog site "*www.dailykos.com*"[1] and 2) "NIPS" which is a corpus of 2,484 scientific papers from the proceedings of NIPS[2]. KOS has $J = 3430$ documents, a vocabulary size of $W = 6909$, a total of $N = 467714$ words, and $M = 360664$ unique word-document pairs. NIPS has $J = 1740$, $W = 12419$, $N = 2166029$ and $M = 836644$. We used $K = 10$ for KOS and $K = 40$ for NIPS and set $\alpha_k = \beta = .1$ for both datasets.

In all sets of experiments, we withheld a random 10% of the words in the corpus for testing, while training on the remaining 90%. We compared a variety of algorithms: collapsed Gibbs sampling (CGS), standard VB (SVB), collapsed VB (CVB), as well as two hybrid algorithms: a hybrid of standard VB and CGS (SVB/CGS) as described in

[1]Downloadable from *http://yarra.ics.uci.edu/kos/*. Thanks to Dave Newman for pointing us to this site.

[2]Originally from *http://books.nips.cc* and preprocessed by Sam Roweis and Dave Newman.

Section 5, and a hybrid of collapsed VB (CVB/CGS) as described in Section 5.1. For both hybrid algorithms we only sampled data-cases for which $\hat{N}_{wj} = 1$. For all algorithms we trained for 300 iterations, testing after every iteration. In the figure captions we report the number of runs over which we averaged the results.

The algorithms were tested using the standard measure of individual word perplexity on the withheld test set. For the pure variational algorithms this is:

$$p(\{x_i^{\text{test}}\}|\{d_i^{\text{test}}\}) = \prod_i \sum_k \frac{\alpha_k + \bar{N}_{kd_i^{\text{test}}}}{\sum_k \alpha_k + \bar{N}_{d_i^{\text{test}}}} \frac{\beta + \bar{N}_{x_i^{\text{test}}k}}{W\beta + \bar{N}_k} \quad (30)$$

For CGS and the hybrid algorithms, we perform an online average over the samples after every iteration of sampling, discarding an initial burn in phase of 10 iterations,

$$p(\{x_i^{\text{test}}\}|\{d_i^{\text{test}}\}) = \prod_i \sum_k \frac{1}{S} \sum_{s=1}^{S} \frac{\alpha_k + \bar{N}^s_{kd_i^{\text{test}}}}{\sum_k \alpha_k + \bar{N}^s_{d_i^{\text{test}}}} \frac{\beta + \bar{N}^s_{x_i^{\text{test}}k}}{W\beta + \bar{N}^s_k} \quad (31)$$

The results for KOS and NIPS are shown in Figures 3, 4, 5 and 6. The variational algorithms converged faster than CGS or the hybrid algorithms, but converged to suboptimal points. Collapsed algorithms performed better than the standard counterparts. The hybrid algorithms significantly improved upon the corresponding pure variational algorithms, with the performance of CVB/CGS being basically on par with CGS. To study how these results depend on the vocabulary size, we first ordered all the word-types according to their total number of occurrences and then only retained the top 3000 most frequent words for KOS and the top 6000 most frequent words for NIPS. The results are shown in Figures 7 and 8. Similar results were obtained with reduced vocabulary sizes of 4000 for KOS and 4000 and 8000 for NIPS. We conclude that for both datasets the perplexities have dropped significantly implying that prediction has become easier. However, the relative performance of the hybrid algorithms has not significantly changed.

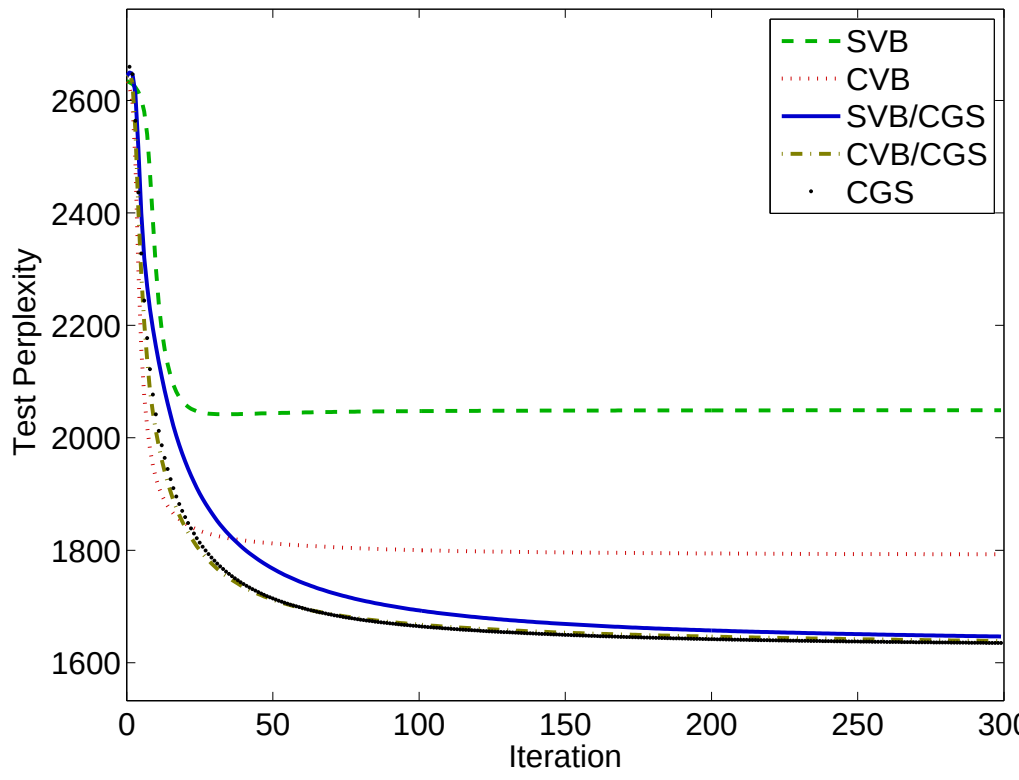

Figure 3: Perplexities of algorithms as function of number of iterations for KOS. Results averaged over 20 runs. The lines for CVB/CGS and CGS are on top of each other.

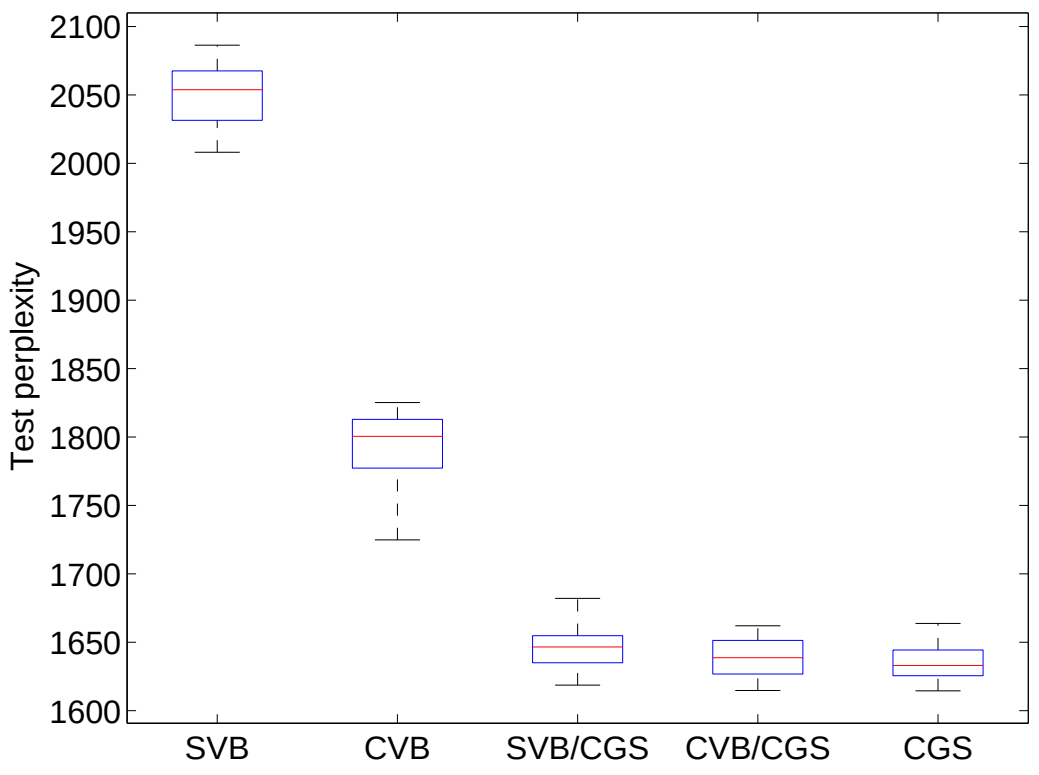

Figure 4: Final perplexities of algorithms at iteration 300 for KOS. Results averaged over 20 runs.

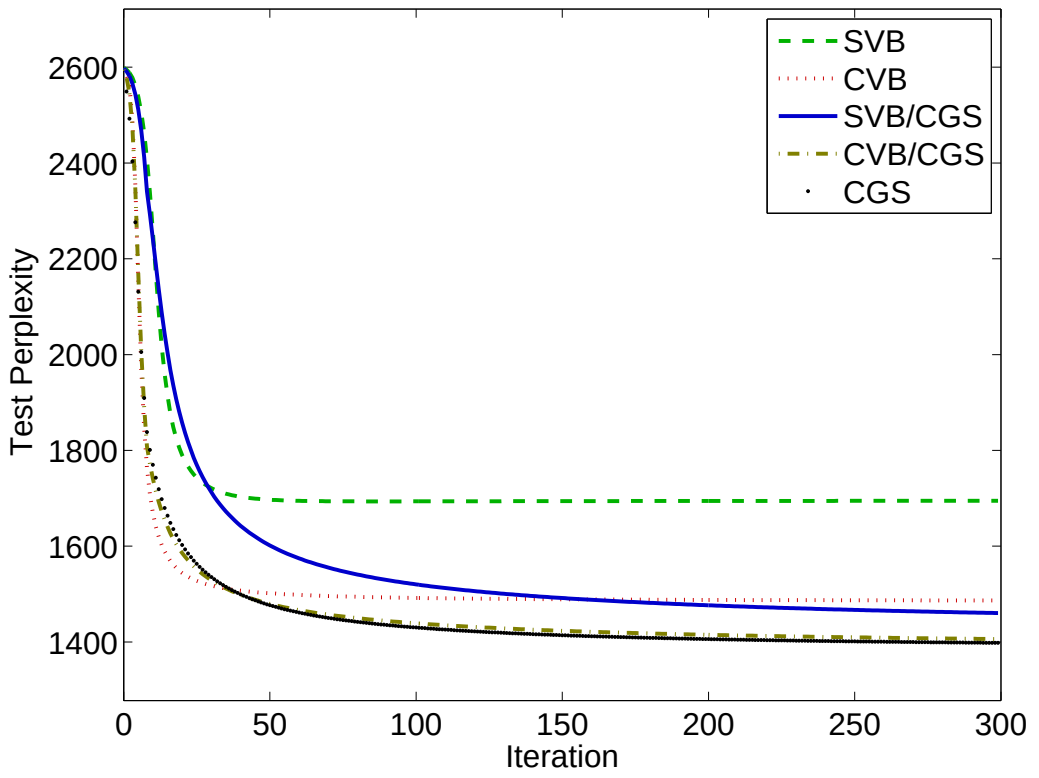

Figure 5: Perplexities of algorithms as function of number of iterations for NIPS. Results averaged over 17 runs.

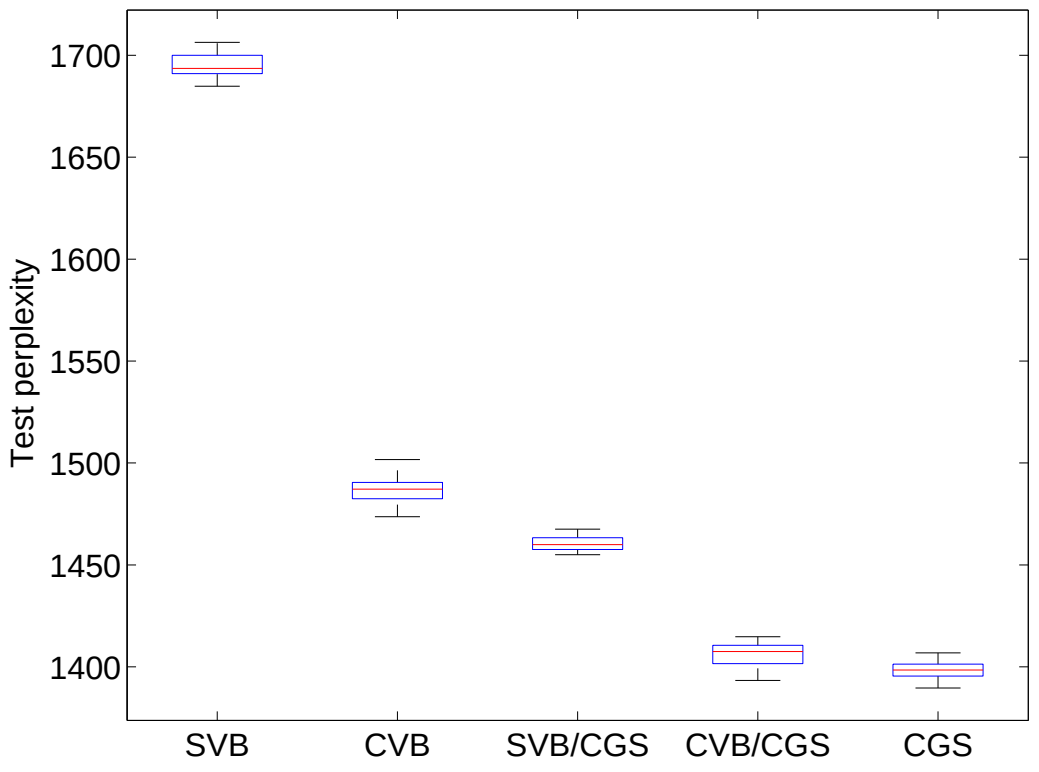

Figure 6: Final perplexities of algorithms at iteration 300 for NIPS. Results averaged over 17 runs.

To understand how much the algorithms learned from the singleton counts, $\hat{N}_{wj} = 1$, we first removed them from the training set but not from test set and subsequently re-moved them from both training and test sets. The results for KOS are shown in Figures 9 and 10 (for a full vocabulary). In this case the difference between hybrid and VB algorithms is only due to the fact that we are still performing an online average for hybrid algorithms, even though we are not sampling any data-cases. We observe that the impact on the absolute values of the perplexities is very large indeed. The results for NIPS were qualitatively similar, but were much more moderate due to the fact that KOS contains many more very small counts than NIPS. We thus conclude that singleton counts play a very important, and sometimes even dominant role for text data in terms of perplexity. Whether this conclusion holds true for actual applications of LDA, such as indexing a corpus or retrieving similar documents to a test document remains to be seen.

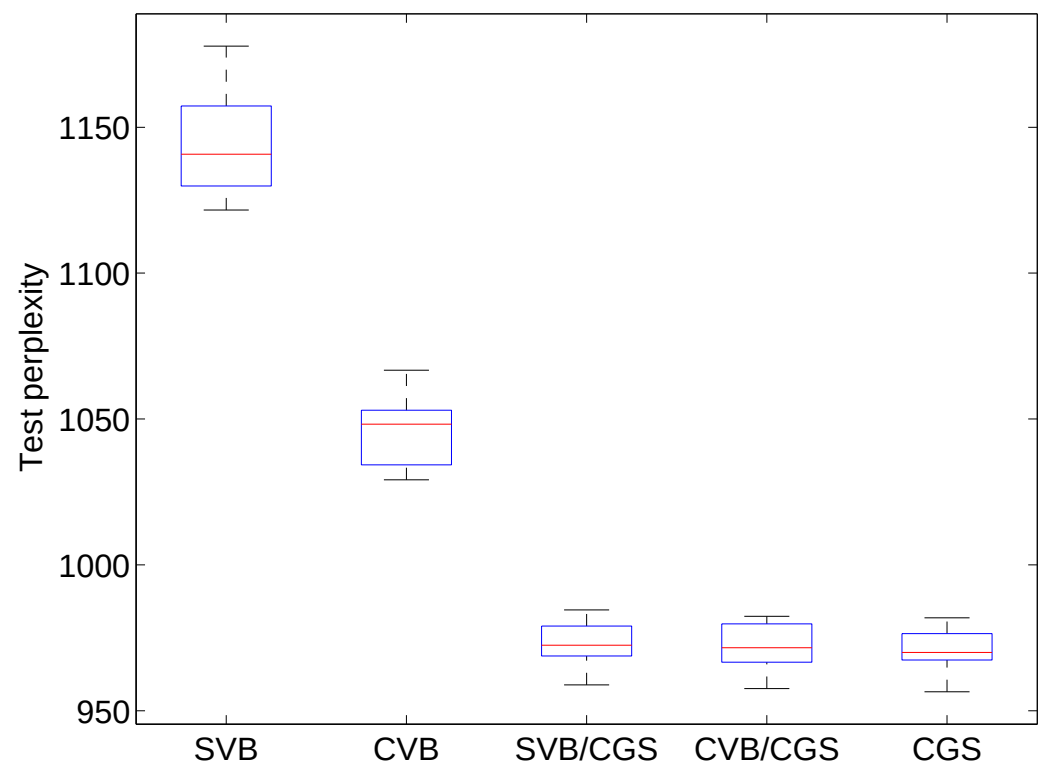

Figure 7: Final perplexities of algorithms at iteration 300 for KOS with a reduced vocabulary size of 3000 word types. Results averaged over 14 runs.

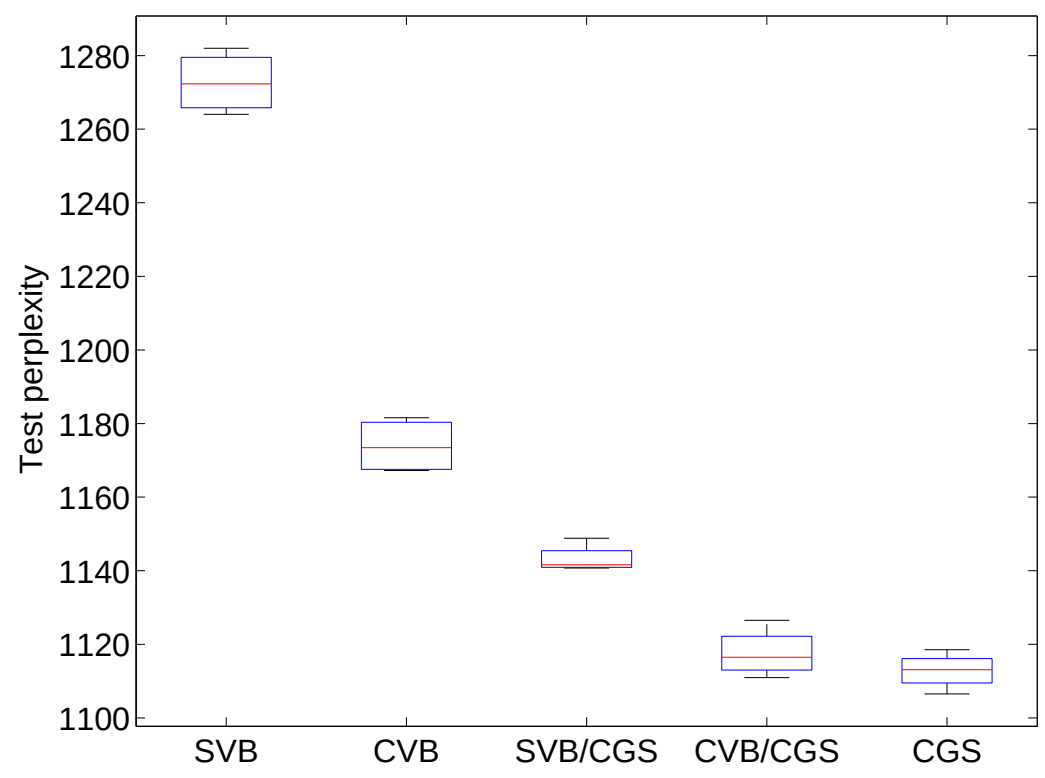

Figure 8: Final perplexities of algorithms at iteration 300 for NIPS with a reduced vocabulary size of 6000. Results averaged over 4 runs.

For text data, where small counts are so dominant VB algorithms are not significantly faster than CGS. This is mainly due to the fact that the VB algorithms must compute[3] $exp(\psi(\cdot))$ which is more expensive than the simple count ratios necessary for CGS. For datasets which have relatively more large counts we expect the VB and hybrid algorithms to be significantly faster than CGS because they require only a single update per nonzero word-document entry rather then $\hat{N}_{wj}$ CGS updates for that entry.

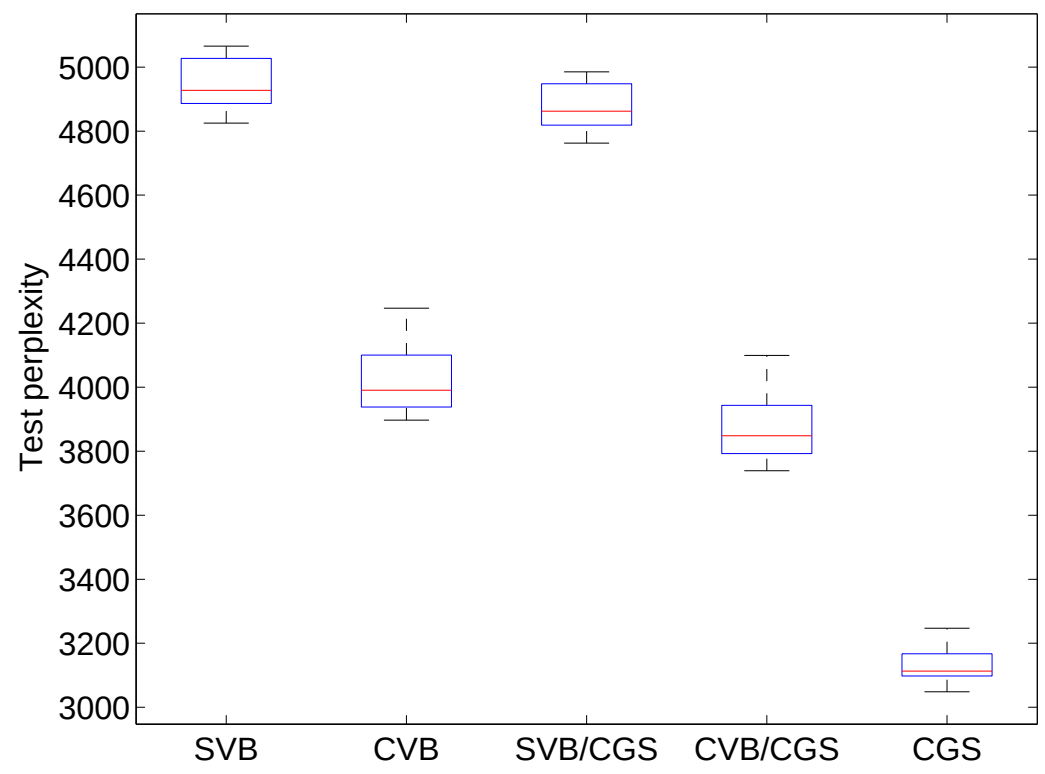

Figure 9: Final perplexities of algorithms at iteration 300 for KOS with all singleton counts removed from only training set. Results averaged over 20 runs.

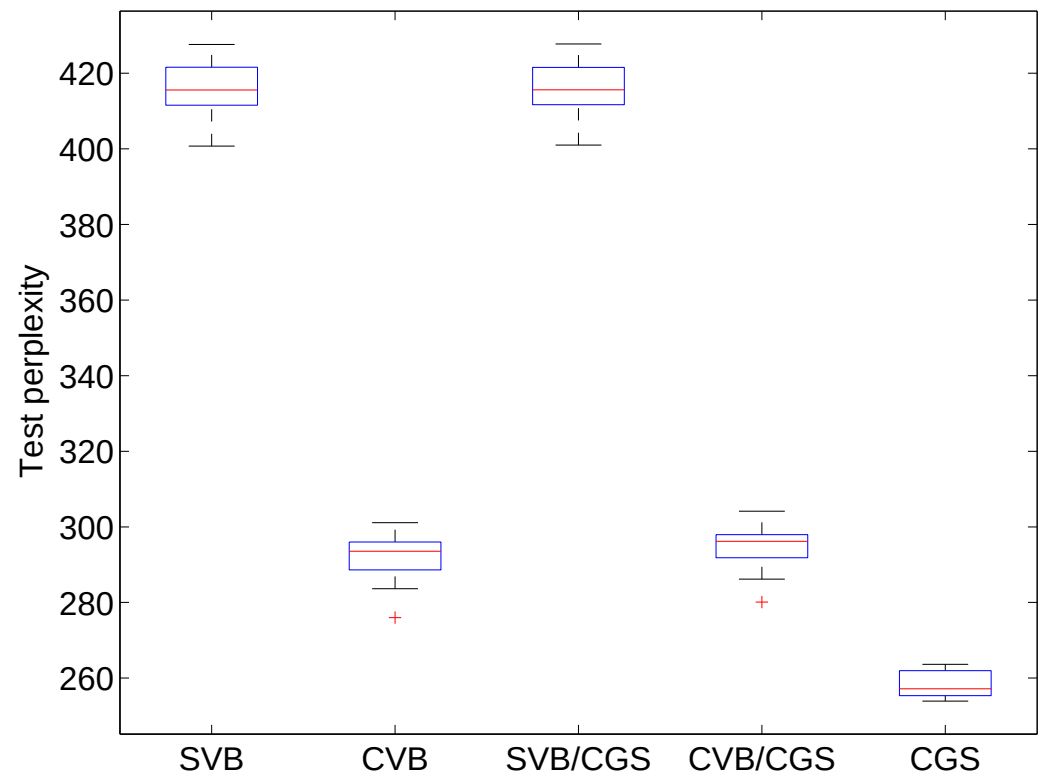

Figure 10: Final perplexities of algorithms at iteration 300 for KOS with all singleton counts removed from both training *and test* sets. Results averaged over 12 runs.

## 8 Discussion and Related Work

In the context of topic models and Bayesian networks, we present a novel hybrid inference algorithm that combines variational inference with Gibbs sampling in a collapsed state space where the parameters have been marginalized out. We split the data-cases into two sets, $\mathcal{S}^{\mathrm{CG}}$, $\mathcal{S}^{\mathrm{VB}}$, where data-cases in the set $\mathcal{S}^{\mathrm{CG}}$ are handled with Gibbs sampling while data-cases in the set $\mathcal{S}^{\mathrm{VB}}$ are handled with variational updates. These updates interact in a consistent manner in that they stochastically optimize a bound on the evidence. In this paper we have restricted attention to discrete models, but extensions to the exponential family seem feasible.

[3] In fact, one could probably speedup the computation by noticing that $exp(\psi(\cdot))$ is almost linear for arguments larger than 5, but we haven't pursued this further.

The algorithm has the same flavor as stochastic EM algorithms where the E-step is replaced with a sample from the posterior distribution. Similarly to SEM (Celeux & Diebolt, 1985), the sequence of updates for $Q_t^{\text{VB}}$ and $\mathbf{z}_t$ for the proposed algorithm forms a Markov chain with a unique equilibrium distribution, so convergence is guaranteed. How different this approximate equilibrium distribution is from the true equilibrium distribution remains to be studied.

The algorithm is also reminiscent of cutset sampling (Bidyuk & Dechter, 2007) where a subset of the nodes of a Bayesian network are sampled while the remainder is handled using belief propagation. It suggest an interesting extension of the proposed algorithm where one specifies a division of both data-cases *and nodes* into subsets $\mathcal{S}^{\text{CG}}$ and $\mathcal{S}^{\text{VB}}$. The nodes in $\mathcal{S}^{\text{VB}}$ should form a forest of low-treewidth junction trees for the algorithm to remain tractable. Alternatively, if the treewidth is too large, one can use loopy belief propagation on the set $\mathcal{S}^{\text{VB}}$. Our current choice for $\mathcal{S}^{\text{CG}}$ and $\mathcal{S}^{\text{VB}}$ was quite naive and motivated by our interest to understand why variational algorithm perform poorly for LDA. However, it seems preferable to develop more principled and perhaps adaptive (online) methods to make this division.

Still other hybridizations between variational and MCMC inference exist (de Freitas et al., 2001; Carbonetto & de Freitas, 2006) where variational distributions are used to guide and improve sampling. However, these algorithms solve a different problem and do not operate in collapsed state spaces.

To conclude, we believe that hybrid algorithms between the two major classes of inference schemes, namely variational and sampling, are a fruitful road to trade off bias, variance, computational efficiency and statistical accuracy.

## Acknowledgements

Max Welling was supported by NSF grants IIS-0535278 and IIS-0447903 and by an NWO (Dutch Science Foundation) “Bezoekersbeurs”. The research reported here is part of the Interactive Collaborative Information Systems (ICIS) project, supported by the Dutch Ministry of Economic Affairs, grant BSIK03024.

## A Lemma-1

**Lemma-1:** The following function is convex as a function of $x$,

$$z(x) = \sum_k \log\Gamma(x_k) - \log\Gamma(\sum_k x_k) \tag{32}$$

**Proof Lemma-1:** We can write z(x) as a log-partition function as follows,

$$Z(x) = \exp(z(x)) = \int \mathrm{d}p_1 p_2..p_K \prod_k p_k^{x_k - 1} = \frac{\prod_k \Gamma(x_k)}{\Gamma(\sum_k x_k)} \tag{33}$$

It follows that the Hessian is equal to the covariance of $\{\log p_k\}$ and hence positive definite.

## References

Beal, M. (2003). *Variational algorithms for approximate bayesian inference* (Technical Report). Gatsby Computational Neuroscience Unit, University College London, London, UK. PhD. Thesis.

Bidyuk, B., & Dechter, R. (2007). Cutset sampling for bayesian networks. *Journal Artificial Intelligence Research*, *28*, 1–48.

Blei, D. M., Ng, A. Y., & Jordan, M. I. (2003). Latent Dirichlet allocation. *Journal of Machine Learning Research*, *3*, 993–1022.

Buntine, W. (Ed.). (2002). *Variational extensions to em and multinomial pca*, vol. 2430 of *Lecture Notes in Computer Science*. Helsinki, Finland: Springer.

Carbonetto, P., & de Freitas, N. (2006). Conditional mean field. *NIPS*.

Castella, G., & Robert, C. (1996). Rao-blackwellisation of sampling schemes. *Biometrika*, *83(1)*, 81–94.

Celeux, G., & Diebolt, J. (1985). The SEM algorithm: a probabilistic teacher algorithm derived from the EM algorithm for the mixture problem. *Comp. Statis. Quaterly*, *2*, 73–82.

de Freitas, N., d. F. R. Hojen-Sorensen, P. A., & Russell, S. J. (2001). Variational mcmc. *UAI '01: Proceedings of the 17th Conference in Uncertainty in Artificial Intelligence* (pp. 120–127).

Griffiths, T., & Steyvers, M. (2002). A probabilistic approach to semantic representation. *Proceedings of the 24th Annual Conference of the Cognitive Science Society*.

Heckerman, D. (1999). A tutorial on learning with bayesian networks. 301–354.

Minka, T. (2000). *Estimating a dirichlet distribution* (Technical Report).

Scott, S. L. (2002). Bayesian methods for hidden Markov models, recursive computing in the 21st century (pp. 337–351. ).

Teh, Y., Jordan, M., Beal, M., & Blei, D. (2004). Hierarchical Dirichlet processes. *NIPS*.

Teh, Y., Newman, D., & Welling, M. (2006). A collapsed variational bayesian inference algorithm for latent dirichlet allocation. *NIPS*.

Teh, Y. W., Kurihara, K., & Welling, M. (2008). Collapsed variational inference for HDP. *Advances in Neural Information Processing Systems*.